\documentclass[10pt,twocolumn,letterpaper]{article}
\pdfoutput=1
\makeindex

\usepackage{cvpr}
\usepackage{times}
\usepackage{epsfig}
\usepackage{graphicx}
\usepackage{amsmath}
\usepackage{amssymb}
\usepackage{epstopdf}
\usepackage{dsfont}
\usepackage{color}

\usepackage[pagebackref=true,breaklinks=true,letterpaper=true,colorlinks,bookmarks=false]{hyperref}
\bgroup\catcode`\#=12\egroup  \newcommand{\rulesep}{\unskip\ \vrule\ }

\cvprfinalcopy 

\def\cvprPaperID{11} 

\ifcvprfinal\fi
\begin{document}

\title{Neural Dataset Generality}


\author{Ragav Venkatesan \\
	Arizona State University\\
	{\tt\small \href{mailto:ragav.venkatesan@asu.edu}{ragav.venkatesan@asu.edu}}
	\and
	Vijetha Gattupalli \\
	Arizona State University\\
	{\tt\small \href{mailto:jgattupa@asu.edu}{jgattupa@asu.edu}}\and
	Baoxin Li \\
	Arizona State University\\
	{\tt\small \href{mailto:baoxin.li@asu.edu}{baoxin.li@asu.edu}}\and
}

\maketitle
\ifcvprfinal\thispagestyle{empty}\fi

\begin{abstract}
	
	Often the filters learned by Convolutional Neural Networks (CNNs) from different datasets appear similar. This is prominent in the first few layers. This similarity of filters is being exploited for the purposes of transfer learning and some studies have been made to analyse such transferability of features. This is also being used as an initialization technique for different tasks in the same dataset or for the same task in similar datasets. Off-the-shelf CNN features have capitalized on this idea to promote their networks as \emph{best transferable} and \emph{most general} and are used in a cavalier manner in day-to-day computer vision tasks. 
	
	It is curious that while the filters learned by these CNNs are related to the \emph{atomic structures} of the images from which they are learnt, all datasets learn similar looking low-level filters. With the understanding that a dataset that contains many such \emph{atomic structures} learn general filters and are therefore useful to initialize other networks with, we propose a way to analyse and quantify generality among datasets from their accuracies on transferred filters. We applied this metric on several popular character recognition, natural image and a medical image dataset, and arrived at some interesting conclusions. On further experimentation we also discovered that particular classes in a dataset themselves are more general than others.
	
\end{abstract}

\section{Introduction}

\begin{figure}[t]
	\begin{center}
		\includegraphics[width=0.99\linewidth]{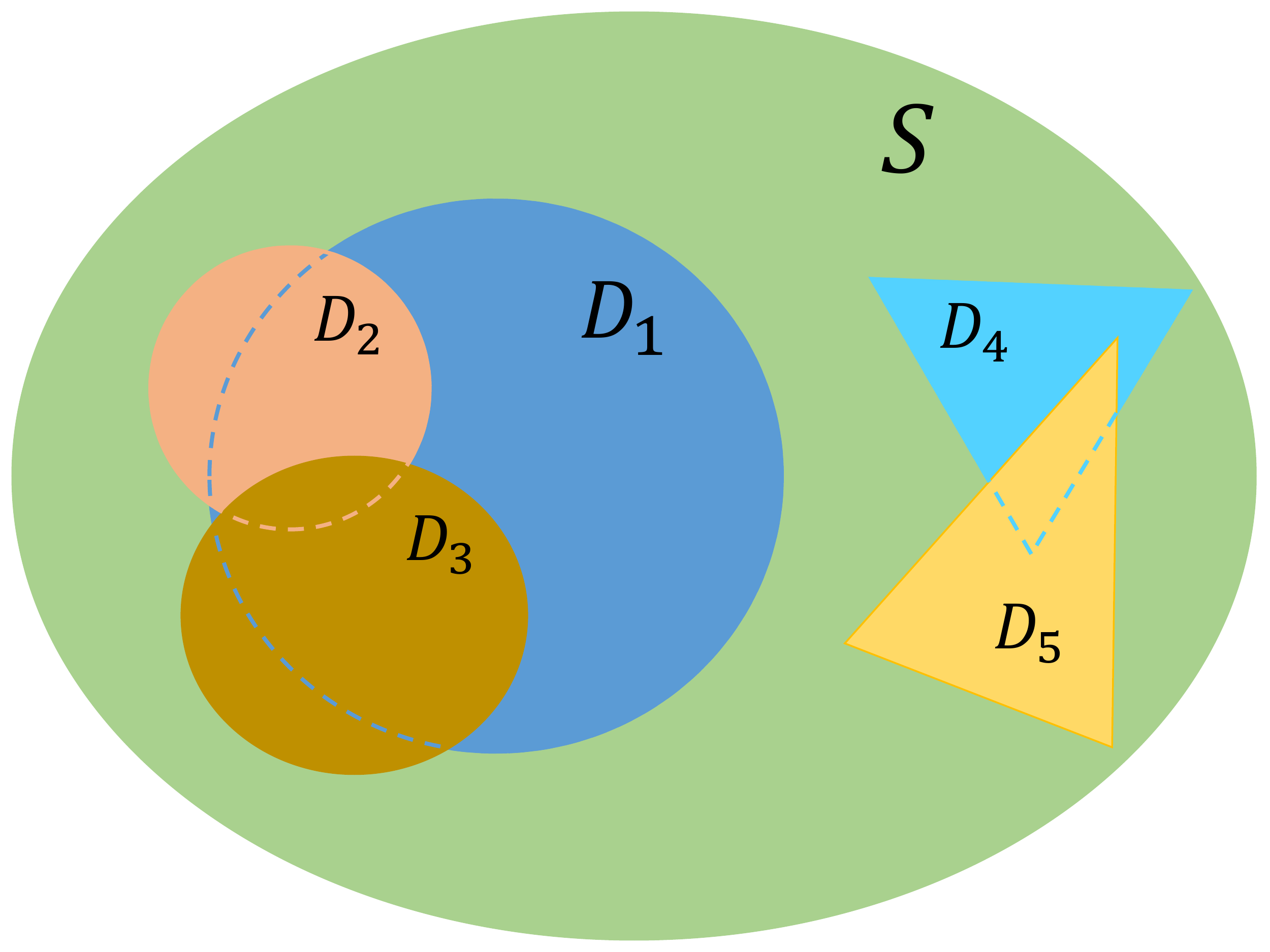}
	\end{center}
	\caption{Thought experiment to describe the dataset generality. $S$ is the space of all possible atomic structures, $D_1 - D_5$ are the atomic structures present in respective datasets.}
	\label{fig:venn}
\end{figure}

Neural networks, particularly CNNs have broken all records recently in the computer vision research area. The growth of CNNs focused initially on the recognition of characters. Fukushima and LeCun were the initial pioneers. Independently they developed CNN based systems, some of which are still being used widely~\cite{fukushima1991handwritten, lecun1989backpropagation}. Large networks are often trained with large number of data samples to achieve good accuracies~\cite{szegedy2014going, krizhevsky2012imagenet}. Still, scepticism over CNNs among the modern day computer vision scientists stems from the fact that one does not have a clear understanding of its inner working. Some studies show that a few ($<1\%$) nodes are all that are actively contributing to classification~\cite{escorcia2015relationship}. They also suggest that large networks often overfit, but since the data is too large over-fitting often works as an advantage~\cite{nguyen2015deep}. 


While it is reasonable to expect edge detectors and Gabor-like features in the lower-level filters and more sophisticated concepts at the higher levels, it is not clear as to why these filters adapt themselves in this manner. What is fairly clear though is that different datasets result in different sets of filters that are similar if the datasets are similar. It is only natural to ask, what role does the data itself play in such filters being learnt and how they compare with filters learnt from another dataset. In this paper we take the view that the filters learnt by networks when trained using a particular dataset represent the detectors for some \emph{atomic structure} in the data itself. In which case each layer is a mapping form the previous layer to the next layer that is constructed using combinations of these atomic structures in the first layer in order to minimize a cost.  

Let us first define \emph{atomic structures} to be the forms that CNN filters take by virtue of the entropy of the dataset it is learning on, analogous to dictionary atoms. Complex datasets have more and varied atomic structures. Consider the following thought experiment: Let's assume that all possible atomic structures reside in an universe $S$. Suppose we have a set of three datasets $ D = \{D_1, D_2, D_3\}$ and $ D \in S $. Consider the system in figure~\ref{fig:venn}. The figure describes the configuration of the elements of $D$. One would now recognize that $D_1$ is a more general dataset with respect to $D_2$ and $D_3$. It is so because, while $D_1$ \emph{contains} most of the atomic structures of $D_2$ and $D_3$, the latter do not contain as many atomic structures of $D_1$. While this analysis is simplified for one layer, in typical CNNs, co-adaptation plays a major role in the learning of these atomic structures. Therefore, generality as defined by the overlap of areas in a layer-wise Venn diagram is impractical to obtain. 

In this paper we postulate that, the generalization performances of CNNs on one dataset re-trained on a network initialized by training using another, could be used to derive generality. We call this process of pre-training as \emph{prejudicing}. By prejudicing on the first dataset, we froze and unfroze layers and retrained the networks on the second dataset. By freezing layers we are making a network more obstinate and we call this process obstination\footnote{Obstinate layer or freezing implies that the weights were not changed during backprop. The layer remains prejudiced.}. The more the layers are frozen, the more obstinate the feature extractor is, therefore the harder the classifier has to work. If the prejudice was general enough, the classifier shall still generalize fairly well enough. What this means is that if the prejudicing dataset is more general than the re-train dataset, the classifier can generalize better than vice versa.

We developed a generality metric by comparing the gain in performances of networks of various obstination. Using a generality such as the one proposed, it becomes clearer as to what kind of datasets are to be used to prejudice CNNs with during transfer learning. We even discovered that samples with particular labels within a dataset alone are general enough. So, if we begin by prejudicing the network on only those and then moved on to the rest of the labels, we were able to learn the rest of the dataset with considerably less training samples while achieving comparable generalization performances. 

Off-the-shelf networks such as VGG, overfeat and various published Caffe model weights are trained on large scale image datasets such as Imagenet or PASCAL~\cite{SimonyanZ14a, jia2014caffe, girshick2014rich, ILSVRC15, everingham2010pascal}. For instance, while these may work on applications such as human pose recognition or vehicle detection, they do not necessarily work on tasks involving medical images. This is because the datasets on which they are trained are not general enough to adapt to the representational requirements of medical images, which is on a manifold unique and disjoint form the manifolds of natural images. This is visualized in $D_4$ and $D_5$ from figure~\ref{fig:venn}. Even a large collection of natural images is not general enough to have networks trained that are suitable to medical images. In these cases, the prejudiced network often fails. For instance, on the Colonoscopy dataset discussed later a $22$ layer deep overfeat features, trained with a logistic regression performs poorer than a $3$ layer deep CNN trained from random initialization, which is in turn outperformed when initialized by a network trained on an endoscopy dataset.

In this article we considered popular offline character recognition datasets and arrived at some interesting analysis and generalities. We also show that within the MNIST dataset, classes $[4,5,8]$ are general enough that we could learn the other classes with very few (even just one) samples, when prejudiced with networks trained on $[4,5,8]$. We also considered more sophisticated datasets such as Cifar 10 and Caltech 101 against some medical image datasets for colonoscopy video quality~\cite{krizhevsky2009learning}. This study led us to two major research insights:
\begin{enumerate}
	\item If one has very few data to learn from, which other dataset is better to prejudice the network with? The answer is particularly helpful when dealing with medical image datasets where data is very scarce and one can't simply use a network trained on VOC datasets as feature extractors as discussed above. 
	\item Among the various classes during the training procedure, if we prejudice with a certain \emph{general} set of classes first and then move on to others later, generalization to all classes, even for those with few samples is better. This is particularly significant if the dataset has a lot of samples in certain classes and not as much of others.
\end{enumerate}

The rest of the paper is organized as follows: section~\ref{sec:related} discusses related works, section~\ref{sec:doe} presents the design of our experiments, section~\ref{sec:results} shows some results on the core-experiment and section~\ref{sec:conclusions} provides concluding remarks.

\section{Related work}
\label{sec:related}

One related work that this article shares with is the work by Yosinski et al~\cite{yosinski2014transferable}. In that article, the authors considered two tasks $A$ and $B$ that were essentially $500$ classes each from the Imagenet dataset~\cite{ILSVRC15}. They trained an $8$ layer network on one of the tasks (say $A$). They then initialized a new network carrying over the first $n$ layers from the previous job while randomly initializing the others. This new network was used to retrain task $B$. Such a network was $AnB^{+}$. They experimented by obstination of the carried over layers. Such a network was $AnB$. They also studied the specificity of each layer and their contributions to the overall performance. They also showed that networks working on similar tasks had a high memorability and that co-adaptation of layers increased the generalization performance. 

While this analysis is interesting, it was performed on only one dataset: Imagenet. By design, the networks were forced to learn very general filters, so as to be best transferable. Since the images were all \emph{natural images}, one would expect the layers to be more Gabor-like at earlier layers and have more label specific features at later layers, which was what was observed. Also, the paper analysed the transferability of the feature extractors from the perspective of the networks in terms of their fall in generalization performance. This analysis was not catered to the dataset's perspective, which is that the filters learned are a property of the dataset being trained on. This was not a problem for the authors as their datasets for tasks $A$ and $B$ occupied similar manifolds. This analysis also didn't explore re-training using the same network but rather went with re-initializing so that they could learn new co-adaptations. This is not interesting to the study of generality as we want to observe the effect of filters transferred from one dataset on another. The more \emph{general} a dataset, the more variety of atomic structures it offers to the network to learn. We used this idea to define a generality metric between two datasets. To do so, we cannot follow the techniques used by Yosinski et al. 

Another closely related work is the work on \emph{dark knowledge} by Hinton et al, ~\cite{hinton2015distilling}. Here the authors suggest that among the various classes in a dataset, there exists some amount of generalization knowledge that could be transferred. The authors construct a large network that learns all its classes. They then go on to train a smaller network with the same dataset (or with a dataset that is missing some of the classes altogether). While training this smaller network though, instead of using the the hard labels, they also use the softmax output from the large network also for backprop. This creates an effect of the larger network \emph{guiding} the smaller network to not just generalize to the dataset, but also to generalize to unseen classes. This is because, as the argument goes, "the network learns the relationship between the classes" and "all the knowledge is among the relative probabilities or softmaxes that the network is almost certain is wrong"~\cite{hinton2015distilling}. 

Although the author retrains an entire network that is randomly initialized using the softmax outputs from a trained network and uses this as prejudice, no information is actually being transferred in terms of actual filters. Ergo, this work, while interesting, also doesn't help in understanding generality of the data itself in a more direct manner.  Some of the claims made by this article though were indirectly and independently verified by us through our generality results.  The basic claim of their work is that among only a handful of classes, there is enough knowledge to generalize to other classes. Unless there exists some generality between classes, training on particular classes will not have been representational enough for the other classes to learn on. We directly verify this by showing that some classes alone have a high generalization to the rest of the dataset and make a similar conclusion from an entirely independent direction of research.

\section{Design of experiments}
\label{sec:doe} 
\begin{figure*}[!t]
	\begin{center}
		\includegraphics[width=0.24\linewidth]{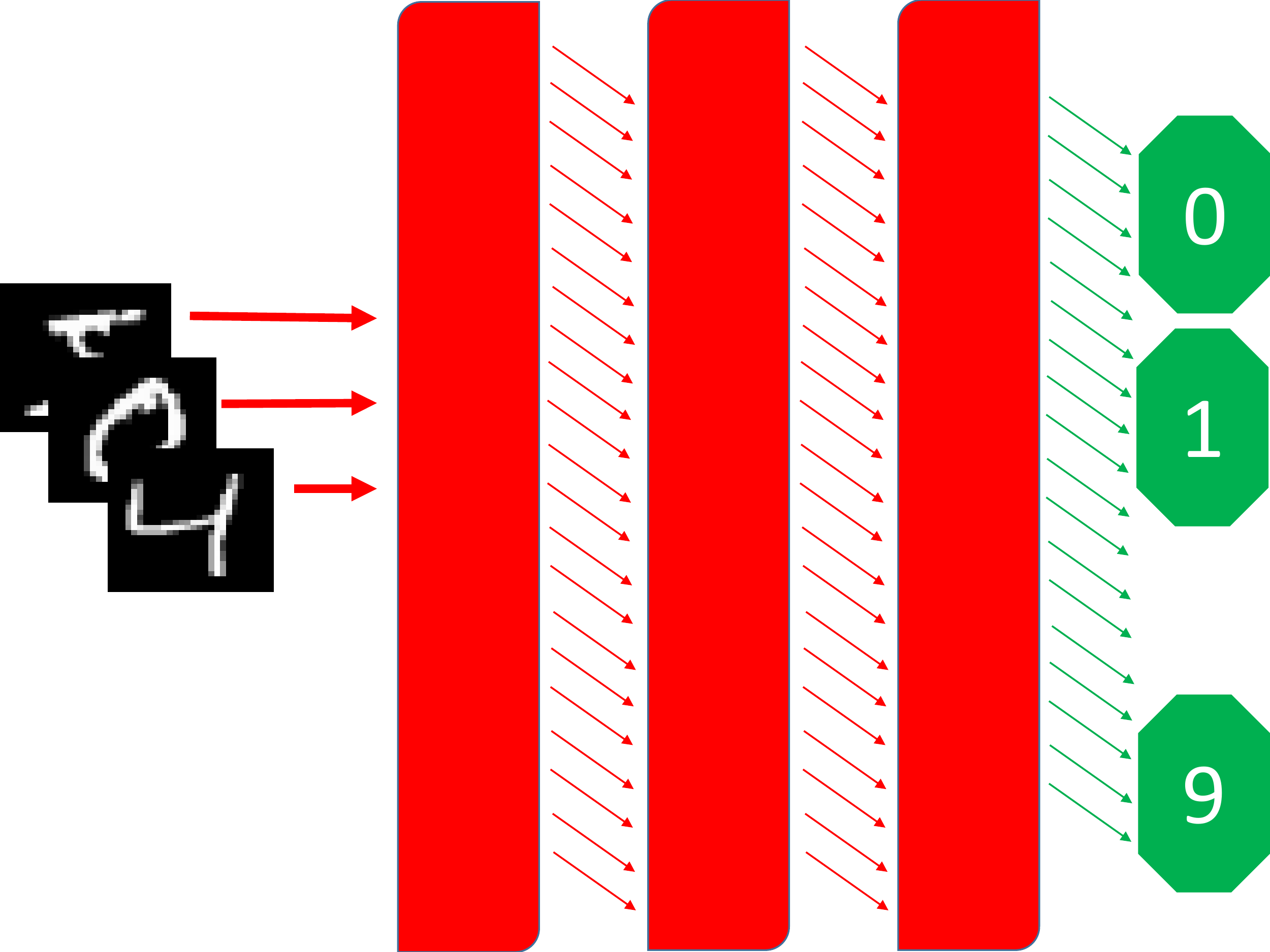} \rulesep
		\includegraphics[width=0.24\linewidth]{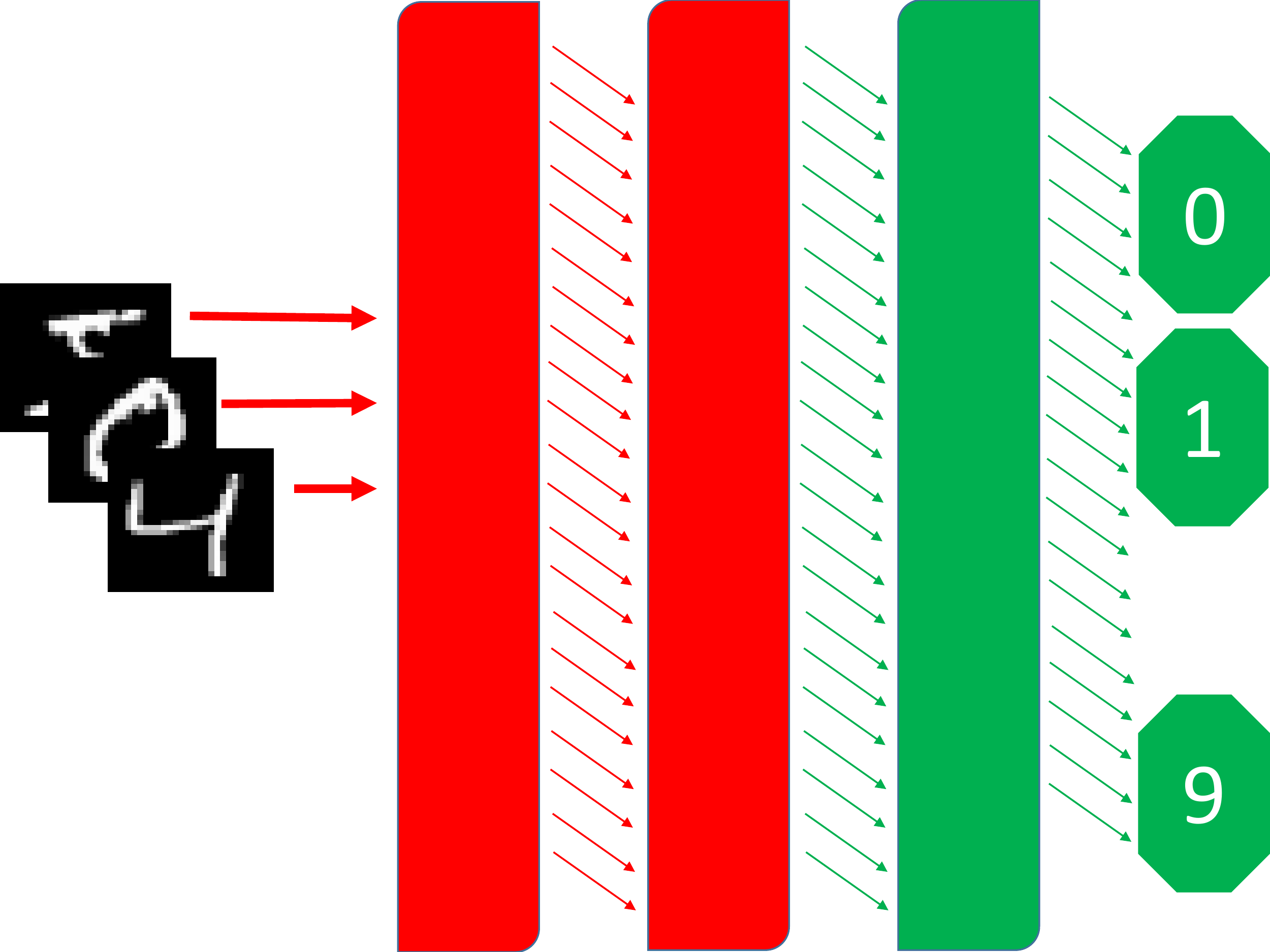} \rulesep
		\includegraphics[width=0.24\linewidth]{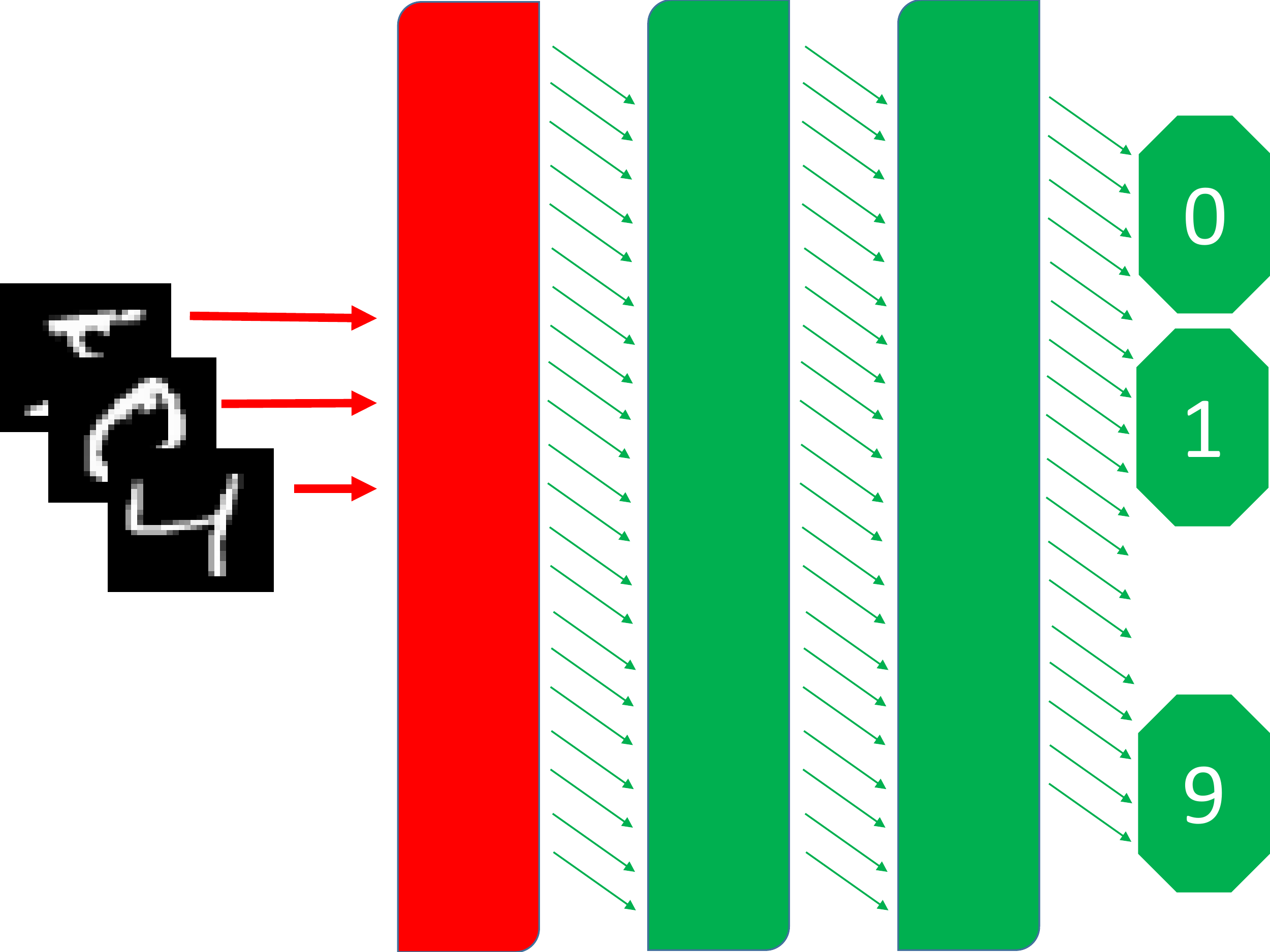}	\rulesep
		\includegraphics[width=0.24\linewidth]{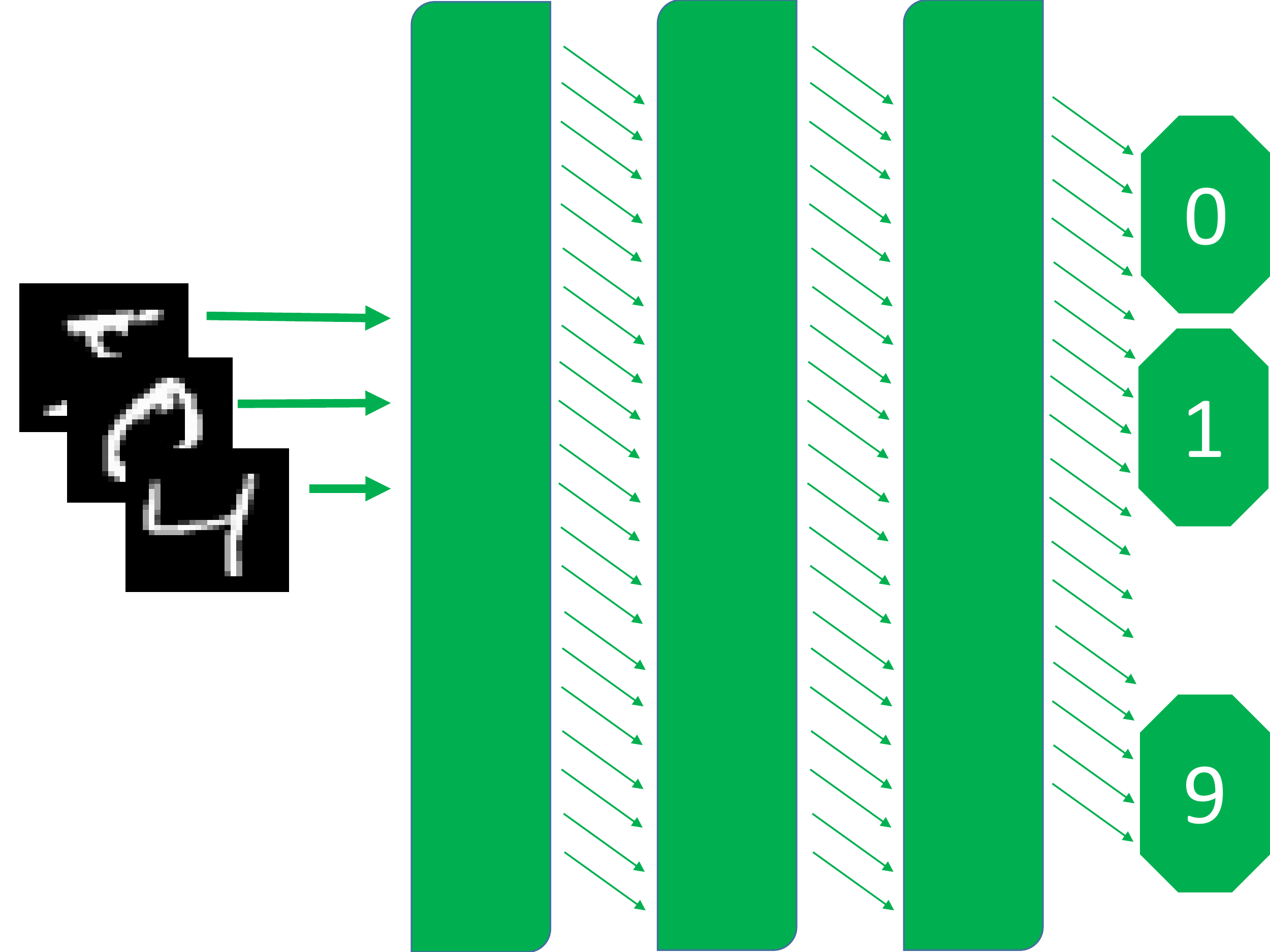} 
	\end{center}
	\caption{Protocol of obstination: From left to right, all layers frozen, one, two and three layers unfrozen. Green represent unfrozen and red represent frozen. Note that the layers are always unfrozen from the end and that the softmax layer is always unfrozen and randomly initialized. This should be generalized similarly for more than three layers also. }
	\label{fig:protocols}
\end{figure*}

\begin{figure}[t]
	\begin{center}
		\includegraphics[width=.99\linewidth]{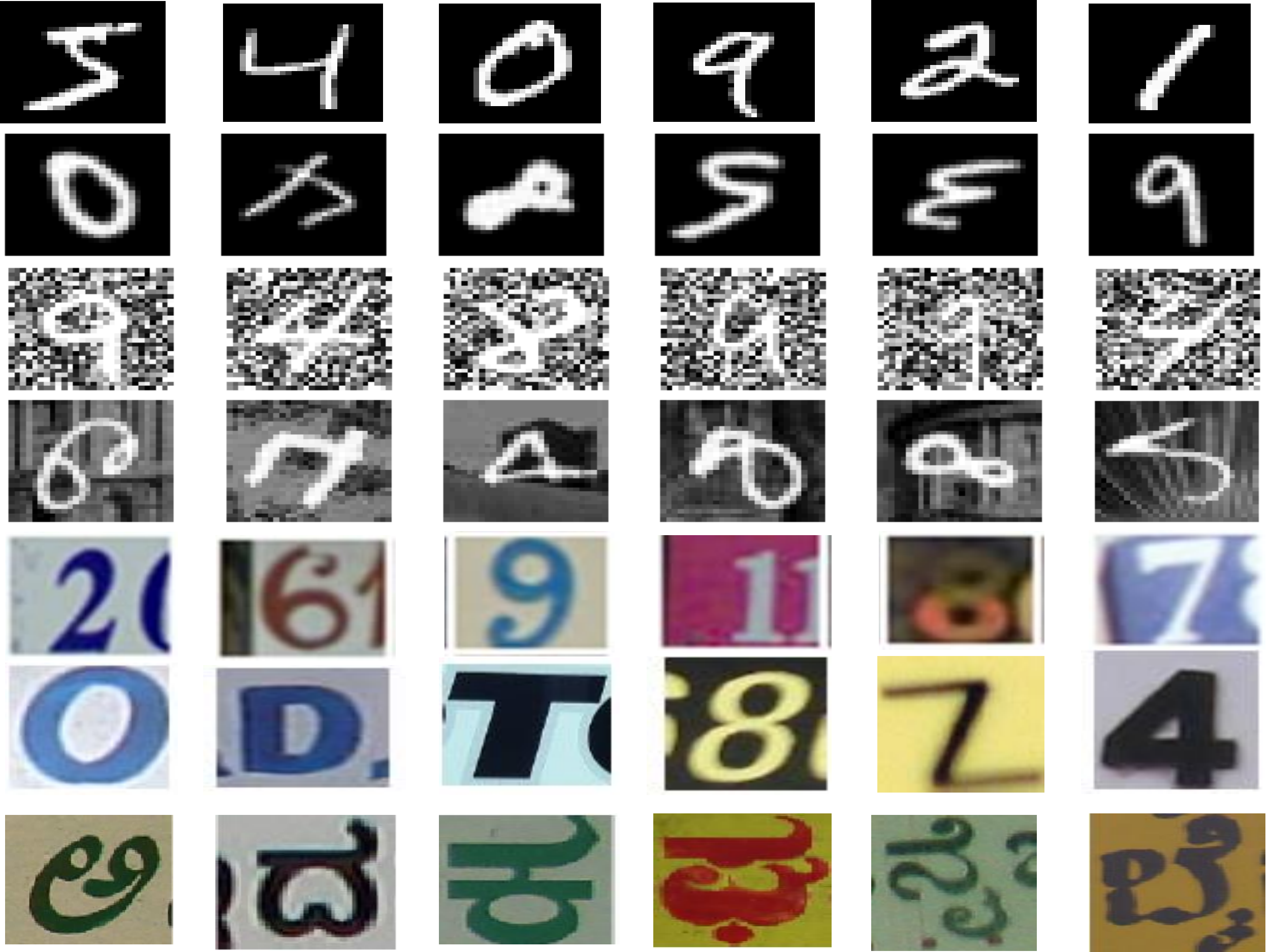}
		\includegraphics[width=0.49\linewidth]{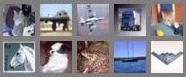}
		\includegraphics[width=0.49\linewidth]{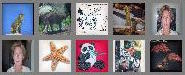}
		\includegraphics[width=0.99\linewidth]{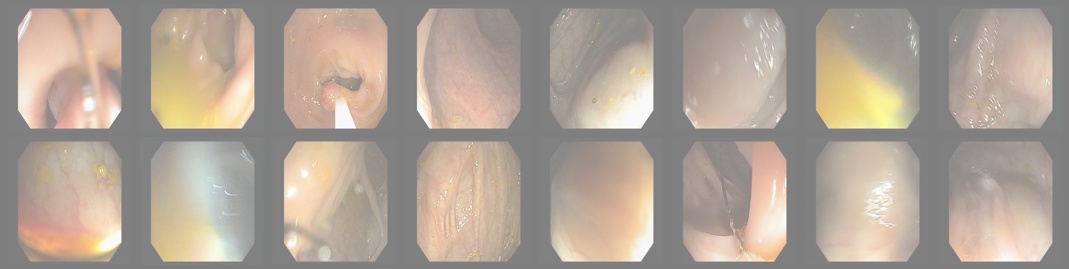}    
	\end{center}
	\caption{Samples of some of the datasets that we used in this analysis. From top to bottom: MNIST~\cite{lecun1998gradient}, MNIST-rotated~\cite{larochelle2007empirical}, MNIST-random-background~\cite{larochelle2007empirical}, MNIST-rotated-background~\cite{larochelle2007empirical}, Google street view house numbers~\cite{netzer2011reading}, Char 74k English~\cite{deCampos09}, Char 74k Kannada~\cite{deCampos09}. Last two rows, first five from left are CIFAR 10 and the rest are Caltech101~\cite{krizhevsky2009learning, fei2007learning}. The bottom row is the colonoscopy dataset.}
	\label{fig:datasets}
\end{figure}

Consider figure~\ref{fig:datasets}. Among the various datasets shown, it is natural to expect any network trained on MNIST to contain simpler filters than MNIST-rotated. This is because, while MNIST-rotated contains many structures from MNIST, due to the rotations, MNIST-rotated will contain additional structures that require the learning of more \emph{complicated} filters. A network trained on MNIST-rotated on its first layers will be expected to additionally have filters for detecting sophisticated oriented edges than for MNIST. This would mean that prejudicing a network with MNIST to then re-train MNIST-rotated is much less helpful than vice versa. A network prejudiced with a general enough dataset is better to be retrained for it generalizes easily. A prejudice must come from a more general dataset if a prejudice transfers positive knowledge as shown in their generalization performances. We use this simple intuition to argue that MNIST-rotated is a more general dataset with respect to MNIST.

Our basic experiment is conducted between pairs of datasets $D_i$ and $D_j$. Firstly, we train (prejudice) a randomly initialized network with dataset $D_i$. We call this network $n(D_i \vert r)$ or the base network ($r$ implies random initialization). We then proceed to retrain $n(D_i \vert r)$ as per any of the setup shown in figure~\ref{fig:protocols}. $n_k(D_j \vert D_i )$ would imply that there are $k$ degrees of freedom, or to be precise, $k$ layers of filters that are allowed to learn by dataset $Dj$ that is prejudiced by the filters of $n(D_i \vert r)$. $n_k(D_j \vert D_i)$ has $N-k$ obstinate layers that carries the prejudice of dataset $D_i$, where $N$ is the total number of layers.  Note that more degrees of freedom implies that the network is less obstinate to learn. Also note that these layers can be both convolutional or fully connected neural layers. Any idea expressed here can be extended to any type of parametrized layers. In fact while we perform operations such as batch normalizations, we even freeze and unfreeze the $\alpha$ and $\beta$ of batch norm~\cite{ioffe2015batch}. Obstination also includes the bias parameters.

Layers learn in two facets. They learn some components that are purely their own and some that are co-adapted from previous layers that are allowed to learn as well. By freezing some layers we are making those layers a fixed functional transformation. Note that the performance gain from $n_k(D_j \vert D_i)$ and $n_{k+1}(D_j \vert D_i)$ is not because of just the new layer $k+1$ being allowed to learn, but of the combination of all $k+1$ layers allowed to learn. 

Figure~\ref{fig:protocols} shows the setup of our experiments and explains degrees of freedom. These are our obstination protocols. Notice that in all the various setup, the softmax layer remains non-obstinate. In fact the softmax layer is always randomly re-initialized because not all dataset pairs have the same number of labels. Also notice that the unfreezing of layers happen from the rear. We cannot unfreeze a layer that feeds into a frozen layer. This is because, while the unfrozen layer learns a new filter and therefore represents the image on new distributed domains, the latter layer is not adapting to such a transformation. When there are two layers unfrozen, the two layers should be able to co-adapt together and must finally feed into an unfrozen classifier layer.

\subsection{Dataset generality}
\label{sec:generality}

Suppose the generalization performance of $n(D_j \vert r)$ is $\Psi(D_j \vert r)$ and the generalization performance of $n_k(D_j \vert D_i)$ is $\Psi_k(D_j \vert D_i)$. First order dataset generality or simply dataset generality of $D_i$ with respect to $D_j$ at the layer $k$ is given by,

\begin{equation}
g_k(D_i,D_j) = \frac{\Psi_k(D_j \vert D_i)}{\Psi(D_j \vert r)}
\end{equation}

This indicates the level of performance that is achieved by $D_j$ using $N - k$ layers worth of prejudice from $D_i$ and $k$ layers worth of features from $D_i$ combined with $k$ layers of novel knowledge from $D_j$ together. Note that the generality is calculated for the base dataset as a measure of how the re-train performs with the prejudice of the base dataset. $g_k(D_i,D_j) > g_k(D_i,D_l)$ indicates that at $k$ layers, $D_i$ provides more general features to $D_j$ than to $D_l$. Conversely, when initialized by $n(D_i \vert r)$, $D_j$ has an advantage in learning than $D_l$.

Note that, $g_k(D_i,D_i) \geq 1 \ \ \forall k $. $g_k(D_i,D_j)$ for $i\neq j$ might or might not be greater than $1$. If $g_k(D_i,D_j) \geq 1$ for $i\neq j$, it indicates that $D_j$ is at least very similar to $D_i$ (such as the case considered by Yosinski et al.) and at most a perfect generalizer of $D_i$~\cite{yosinski2014transferable}. 

\subsection{Class generality}
\label{sec:class-gen}
$D_i$ and $D_j$ need not be entire datasets but can also be just disjoint class instances of the same dataset that is split in two. These generalities will tell us if particular classes are themselves more general than others. For instance, we divided the MNIST dataset into two parts. The first part contained the classes $[4,5,8]$, the rest were contained by the second part\footnote{We chose this combination of classes strategically after trail and error as these are the most general among the classes and exaggerate the effect.}. We performed the generality experiments with MNIST$[4,5,8]$ as base, which was trained over a random initialization. We re-trained this prejudiced network using the second part with the same experiment design as above. We defined class generality as the generality, of a class or a set of classes, retrained on the prejudice of the other mutually exclusive classes. 

We repeated this experiment several times with decreasing number of training samples per-class in the retrain dataset of MNIST $[0,1,2,3,6,7,9]$. All the while, the testing set remained the same size. This implies that the prejudiced network retrains on a much smaller dataset and tests on a much larger dataset. The re-train dataset had $7$ classes. We created seven such datasets with $7p, \ p \in [1,3,5,10,20,30,50]$ samples each. We now define sub-class generality as the generality of these sub-sampled datasets (in each class we only consider a small random sample), retrained on the base of other mutually exclusive classes (MNIST$[4,5,8]$). . Initializing a network that was trained on only a small sub-set of well-chosen classes can significantly improve generalization performance on all classes, even if trained with arbitrarily few samples, even at the extreme case of one-shot learning.

\subsection{Datasets Used}

We designed these experiments across three board categories of datasets: 1. Character datasets that included MNIST~\cite{lecun1998gradient}, MNIST-rotated~\cite{larochelle2007empirical}, MNIST-random-background~\cite{larochelle2007empirical}, MNIST-rotated-background~\cite{larochelle2007empirical}, Google street view house numbers~\cite{netzer2011reading}, Char 74k English~\cite{deCampos09} and Char 74k Kannada~\cite{deCampos09} 2. Natural image datasets that includes Cifar 10 and Caltech 101~\cite{krizhevsky2009learning, fei2007learning} and 3. Natural images against medical images that included in addition to Caltech 101 a Colonoscopy video qualitty dataset. We leave it to the reader to find for themselves details about the datasets from the original articles, but the setup we have used can be found in table~\ref{table:dataset}.  Although we chose only a handful of datasets, the intention of this article was only to show that such generality measures could be made. The scope of this article was not to benchmark various publicly available popular datasets. Neither was it to make suggestions specific to types of datasets.

\subsection{Network architecture and learning}

We used one standard network architecture for all character datasets and experiments, one for Cifar 10 vs. Caltech 101 and another standard for Caltech 101 vs. Colonoscopy. The setup we have used can be found in table~\ref{table:dataset}.

\begin{table*}[t]
	\begin{center}
		\begin{tabular}{||c|| c | c | c | c | c ||}
			\hline
			Dataset 							& 	Training 	& 	Testing 	& 	Validation	& Classes	& Training Batch Size 	\\	
			\hline \hline	
			MNIST ~\cite{lecun1998gradient}		&	50,000 		& 	10,000		&	10,000		&  10		&		500				\\
			MNIST-random-background
			~\cite{larochelle2007empirical}		&  40,000 		& 	12,000		& 	10,000		&	10		&		500				\\
			MNIST-rotated-background
			~\cite{larochelle2007empirical}		&  40,000 		& 	12,000		& 	10,000		&	10		&		500				\\
			\hline	
			NIST Special Dataset-19~\cite{nist}	&	271,220 	& 	271,220		&	271,220		&	62		&		191				\\
			Google Street View House Numbers
			~\cite{netzer2011reading}			&	63,042		&	63,042		& 	63,042		&	10		&		399				\\
			\hline
			Char 74k English~\cite{deCampos09}	&	9,300		&	3,355		&	305 		&	62		&		305				\\
			Char 74k Kannada~\cite{deCampos09} 	&	5,694		&	1,314		&	1,752		&  100		&		438				\\
			\hline \hline	
			MNIST $[4, 5, 8]$					&	14,000 		&	2,500		&	2,500		& 	3		&		500			 \\																										
			MNIST $[0,1,2,3,6,7,9] - p$ per-class 		&	7$p$ 	&	7,000 		&	7,000   & 	7	    &       500          \\	
			\hline \hline
			CIFAR 10~\cite{krizhevsky2009learning}&  40,000	    &   10,000		&   10,000		&  10       &       500          \\
			Caltech 101~\cite{fei2007learning}  &     5,080     &    3,048      &    1,016      &  102      &       254          \\
			\hline
			Colonoscopy\footnote{http:
				//www.polyp2015.com/wp/}            &      2,700    &    900        &     100       &   2       &       100           \\

			\hline
			\hline 
		\end{tabular}
	\end{center}
	\caption{Datasets used and their properties.}
	\label{table:dataset}
\end{table*}

\begin{figure*}[!ht]
	\begin{center}

		\includegraphics[width=0.495\linewidth]{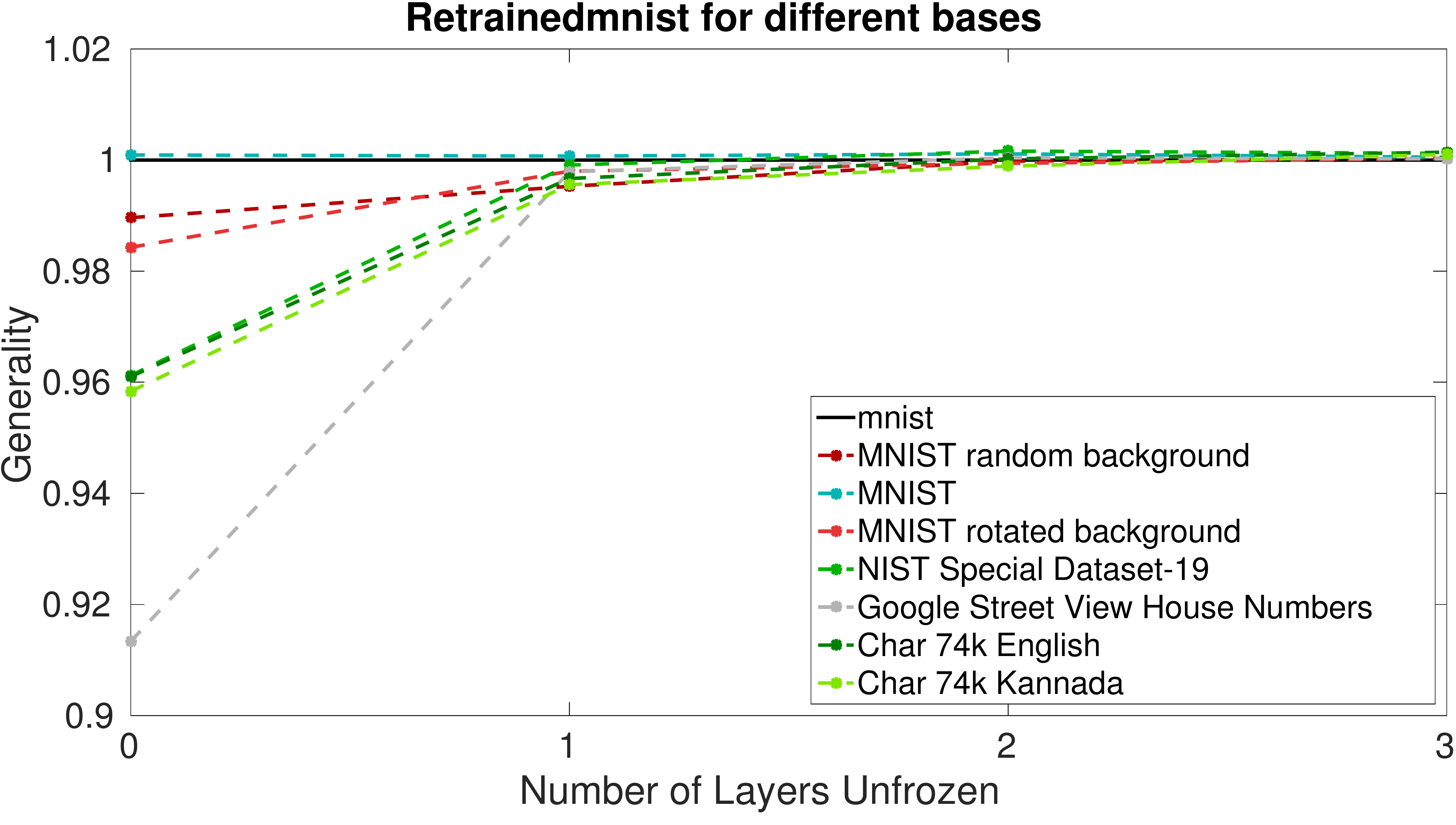} 
		\includegraphics[width=0.495\linewidth]{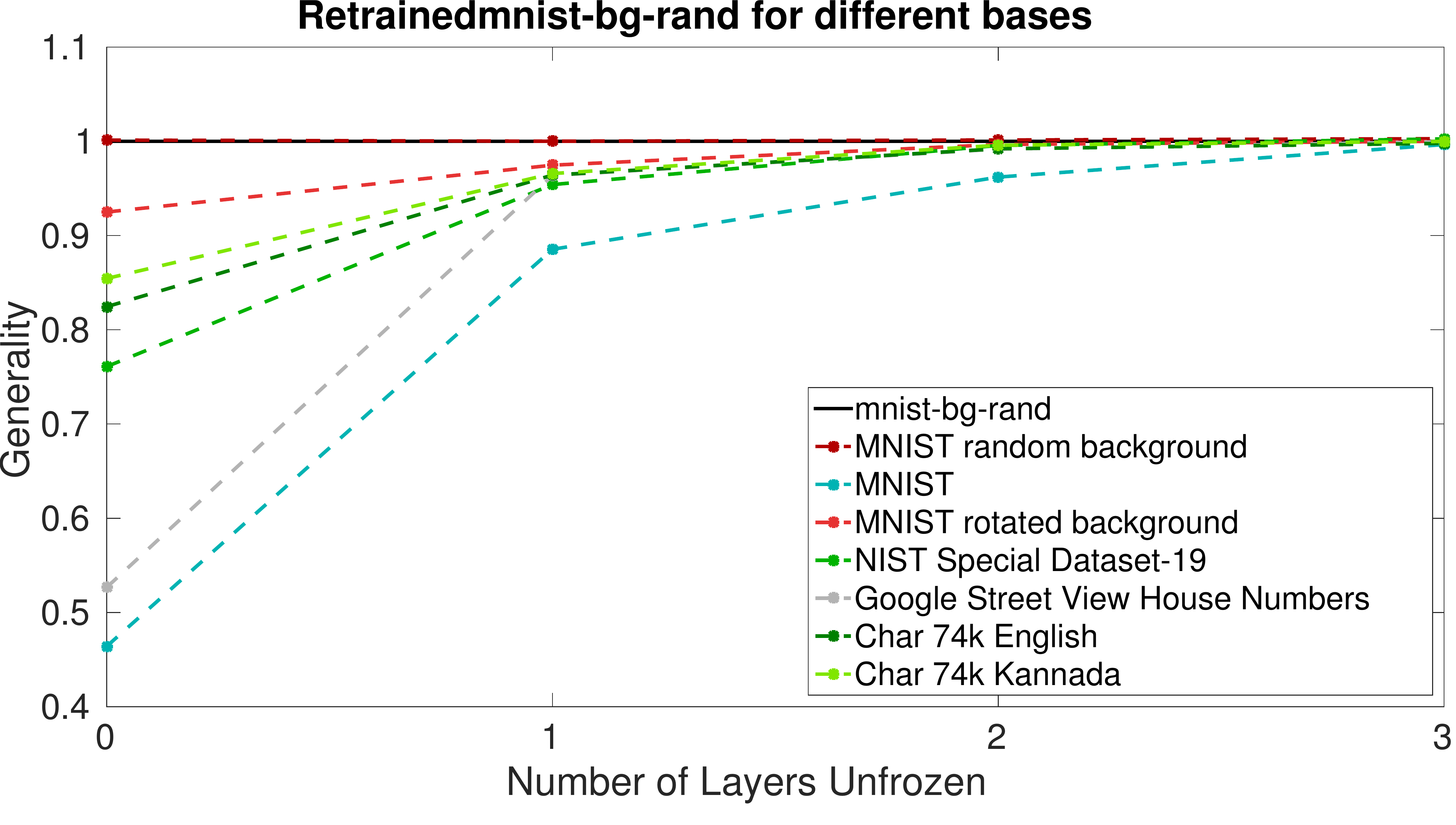} 
		\includegraphics[width=0.495\linewidth]{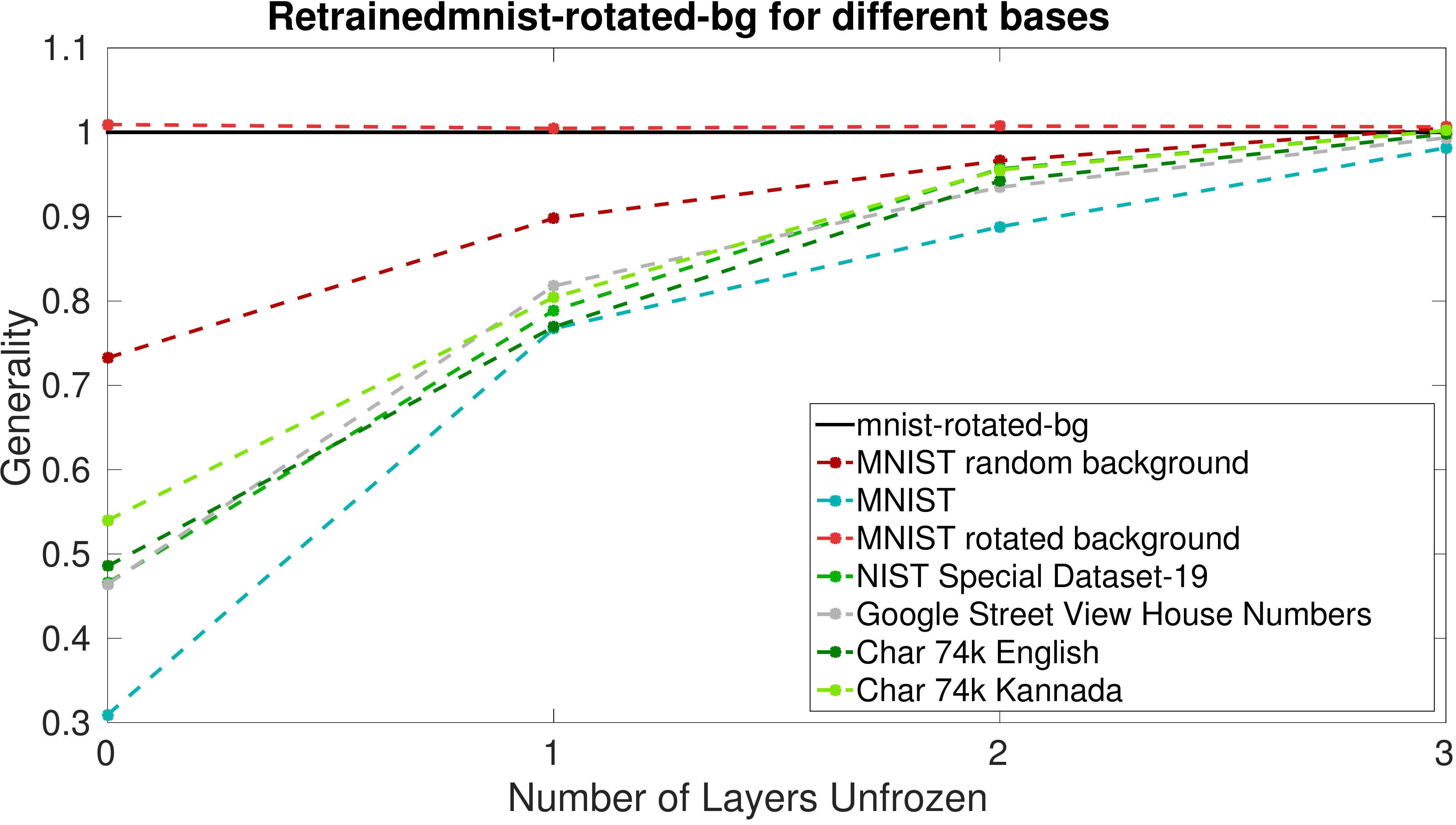} 
		\includegraphics[width=0.495\linewidth]{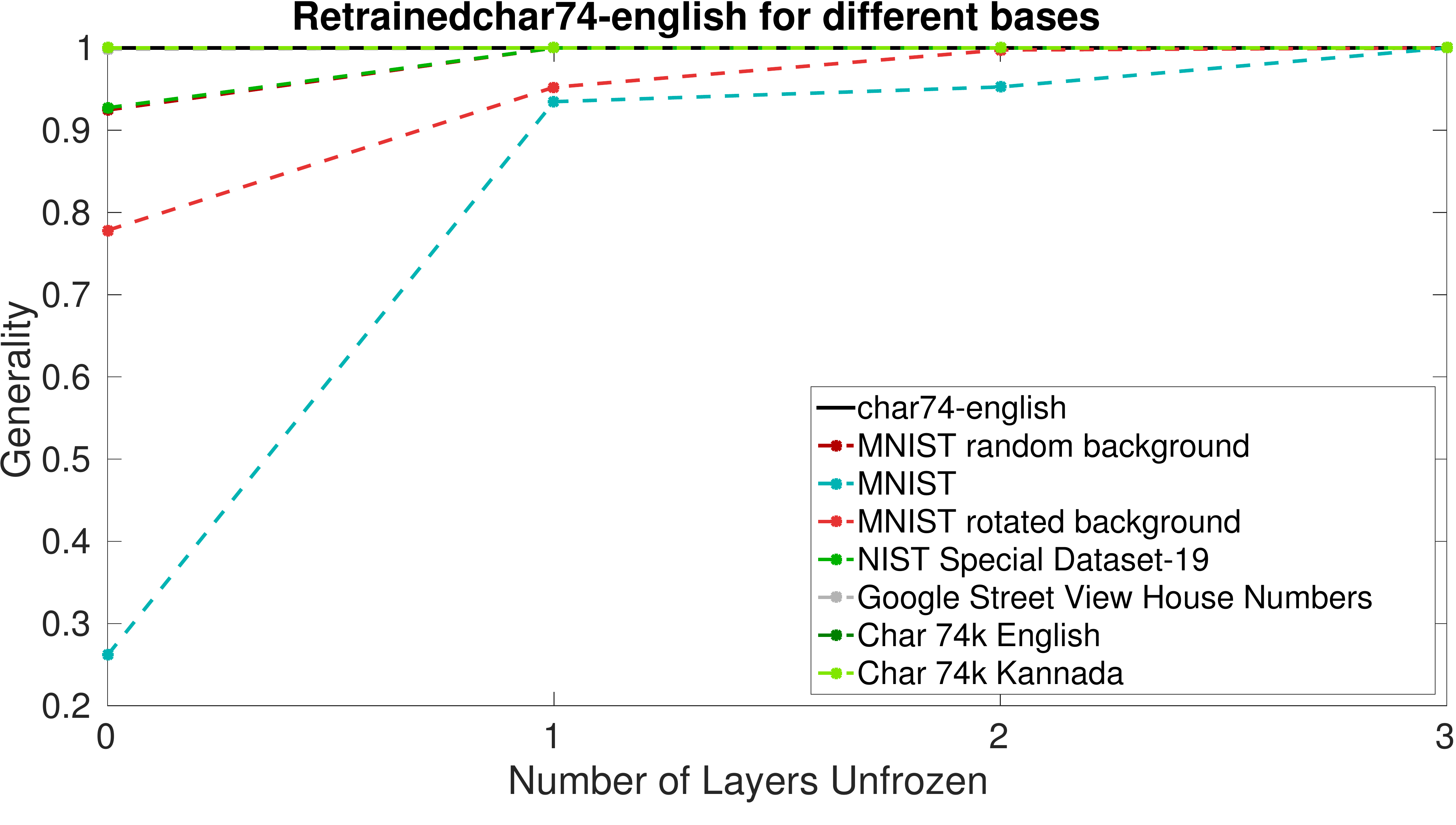}
		\includegraphics[width=0.49\linewidth]{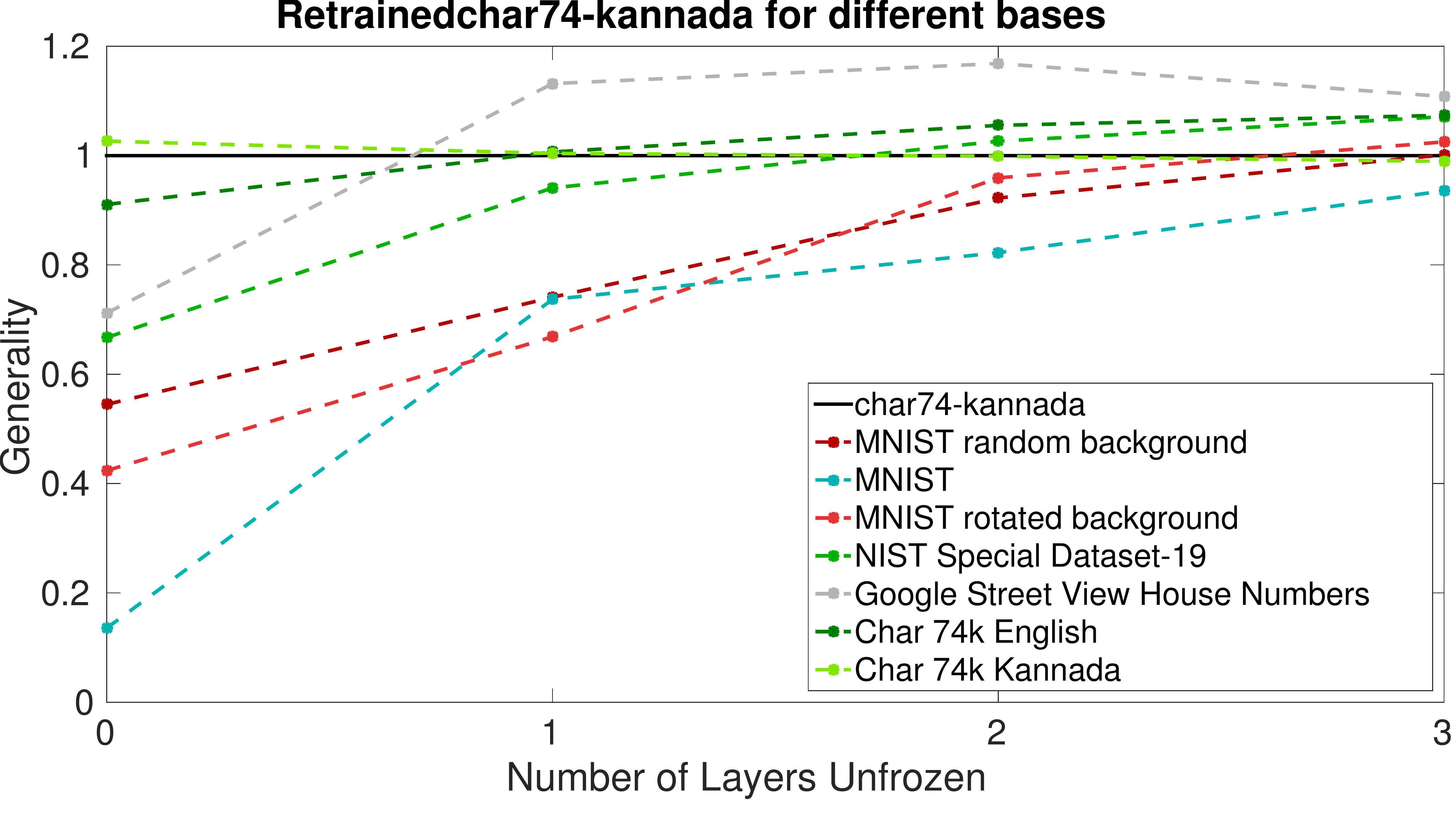} 
		\includegraphics[width=0.495\linewidth]{english.pdf} 
		\includegraphics[width=0.49\linewidth]{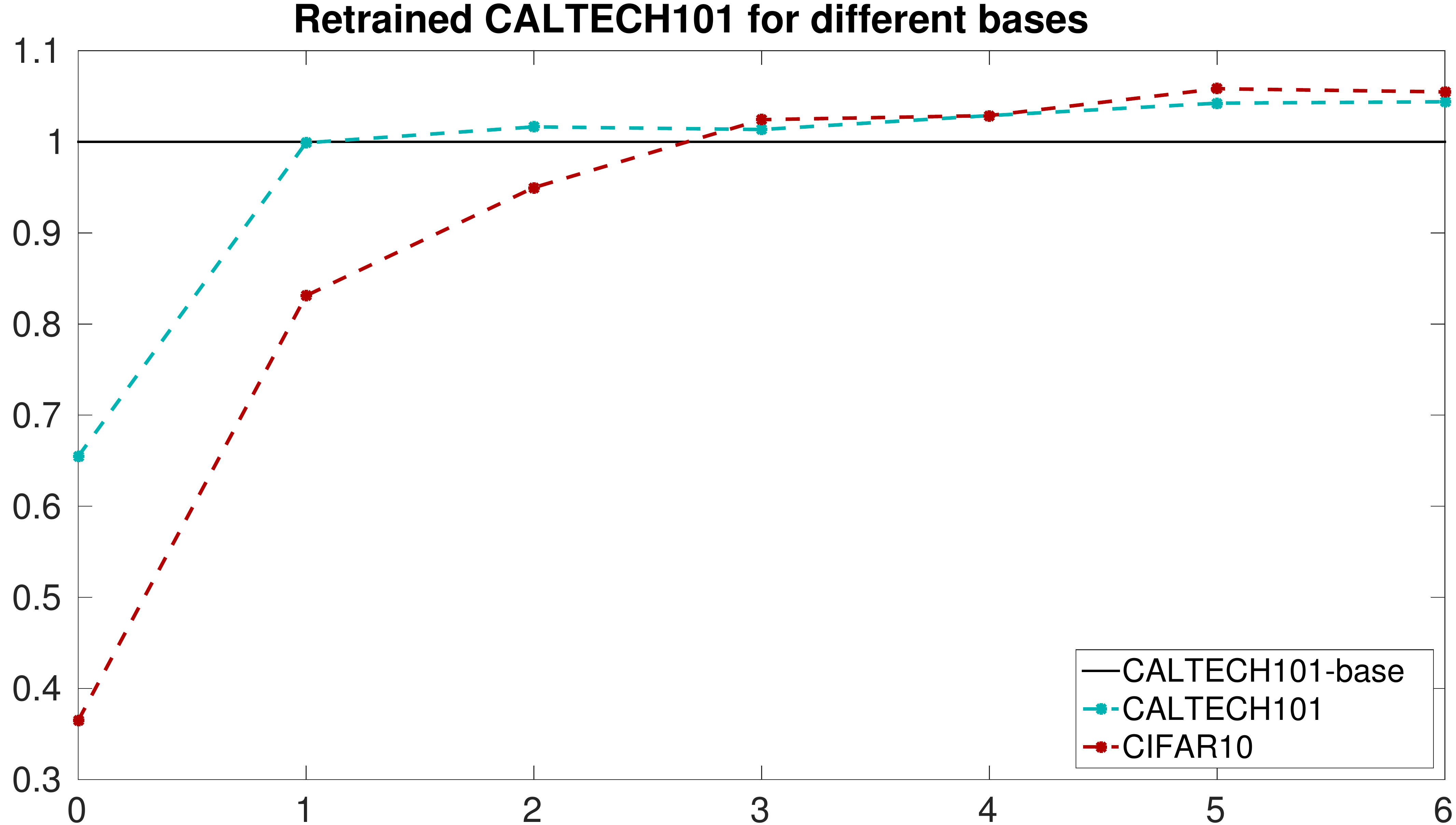} 	    				 
	\includegraphics[width=0.495\linewidth]{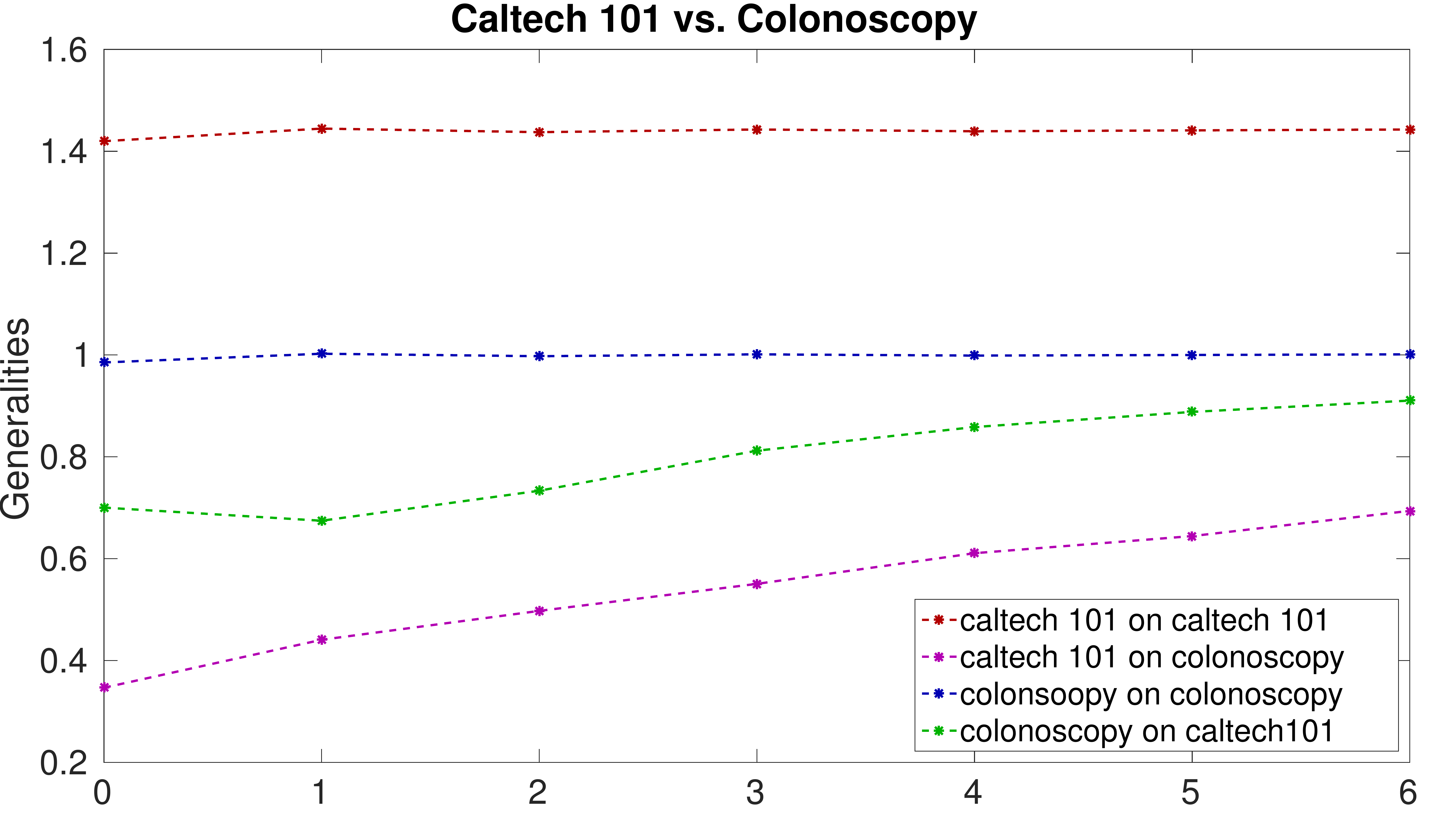} 				
	\end{center}
	\caption{Generalities of datasets not shown in the actual paper. The dark line represents the accuracy of $n(D\vert r)$. Please zoom on a computer monitor for closer inspection.}
	\label{fig:gen-results}
\end{figure*}

The network architectures, learning rates and other details are provided below. The experiments were conducted on a Macbook Pro Laptop using an Nvidia GT 750M GPU, for character datasets and on an Nvidia Tesla K40 GPU for the others, with \href{https://developer.nvidia.com/cudnn}{cuDNN} v3 and \href{http://www.nvidia.com/object/cuda_home_new.html}{Nvidia CUDA} v7. 

Table~\ref{table:dataset} shows the train-test-validation splits and the batch sizes used in stochastic gradient descent of all the datasets used. No pre-processing were done on the images themselves except for cropping, resizing, normalizing. The images were all normalized to lie in $[0,1]$. The character recognition datasets were all of a constant $28X28$ grayscale, the Caltech 101 vs. Cifar 10 experiments were performed ar $32X32$, RGB and the Caltech 101 vs. Colonsoscopy were at $128X128$, RGB. It is to be noted that the aim of the authors was not to set up the networks to achieve state-of-the-art. The authors did although try to achieve satisfactory performances on all base datasets involved before proceeding with the experimentation. 

\subsubsection*{Character Datasets.}
Our networks had three convolutional layers with $20, 20$ and $50$ kernels respectively. All the filters were $5$ X $5$ and were all stride $1$ convolutions. The first layer didn't have any pooling. The second and the third layer maxpool by $2$ subsampled. All the layers used rectified linear units ($ReLU$) activations ~\cite{nair2010rectified}. The classifier layer was a softmax layer and we didn't use any fully connected layers. We used a dropout of $0.5$ only from the last convolutional layer to the softmax layer~\cite{srivastava2014dropout}. We optimized a categorical cross-entropy loss using an rmsprop gradient descent algorithm~\cite{dauphin2015rmsprop}. For acceleration we used Polyak Momentum that linearly increases in range $[0.5, 1]$ from start to $100$ epochs~\cite{polyak1964some}. Unless early terminated, we ran $200$ epochs. We also used a constant $L_1$ and $L_2$ regularizer co-efficients of $0.0001$. Our learning rate was a $0.01$ with a multiplicative decay of $0.0998$. 

\subsubsection*{CIFAR10 Vs. Caltech101 and Caltech 101 vs Colonoscopy.}
For this task, the networks had five convolutional layers with $20, 20$, $50$, $50$ and $50$ kernels respectively. We also had a last fully connected layer of $1800$ nodes, which also had a dropout of $0.5$. All the filters were $5$ X $5$ and were all stride $1$ convolutions. Only the last layer maxpool by $2$ subsampled. All the layers used rectified linear units ($ReLU$) activations ~\cite{nair2010rectified}. All CNN and MLP layers were also batch normalized~\cite{ioffe2015batch}.The classifier layer was a softmax layer and we didn't use any fully connected layers. We used a dropout of $0.5$ only from the last convolutional layer to the softmax layer~\cite{srivastava2014dropout}. We optimized a categorical cross-entropy loss using an rmsprop gradient descent algorithm~\cite{dauphin2015rmsprop}. For acceleration we used Polyak Momentum that linearly increases in range $[0.5, 0.85]$ from start to $100$ epochs~\cite{polyak1964some}. We use a learning rate of $0.001$ for the first $150$ epochs and then fine tune with a learning rate of $0.0001$ for an additional $50$ epochs unless early-terminated. Our learning decay of was subtractive $0.0005$. Figure~\ref{fig:gen-results} shows more generality curves.

\section{Results and observations}
\label{sec:results}  

\subsection{Character Datasets}
Figure~\ref{fig:gen-results} shows the generalities of MNIST-rotated-bg and Kannada prejudiced by all other the character datasets. For reference each plot also shows the generalization performance of a randomly initialized base convolutional network. The following are some observations of interest: 

While no dataset is qualitatively the most general, it is quite clear that \emph{MNIST dataset is the most specific}. Rather, MNIST dataset is one that is generalized by all datasets very highly at all layers. Surprisingly, MNIST dataset actually gives better accuracy when prejudiced with other datasets, rather than when initialized with random, if all layers were allowed to learn. This is a strong indicator that \emph{all datasets contain all atomic structures of MNIST}.  

\emph{NIST, Char74-English and Char74-Kannada follow similar generalization trends with almost all the datasets}. With no degrees of freedom they all generalize rather poorly, but their generalities shoot up once one or many layers of the base networks are unfrozen. This indicates two properties: Firstly, these three datasets have similar manifolds. Secondly this also indicates that the last layers of the base datasets are extracting some particular quality of atomic structures that are present in the these datasets alone. Similarly, SVHN does not generalize in the first layer to most datasets, it generalizes much better in the latter layers. This is particularly noticeable in MNIST and Kannada. This further exemplifies the results. 

While initially one would have assumed that Kannada would be a general dataset, we observed the contrary. SVHN, Char74-English and Nist generalizes better to Kannada than even Kannada itself does. \emph{English characters seem to be a more general set than Kananda.} While counter-intuitive, this result is immediately obvious when one pays close attention to the filers that are learnt and the dataset itself. Kannada is dominated by predominantly curved edges only, whereas even MNIST has a multitude of unique atomic structures.

\begin{figure}[!t]
	\begin{center}
		\includegraphics[width=0.99\linewidth]{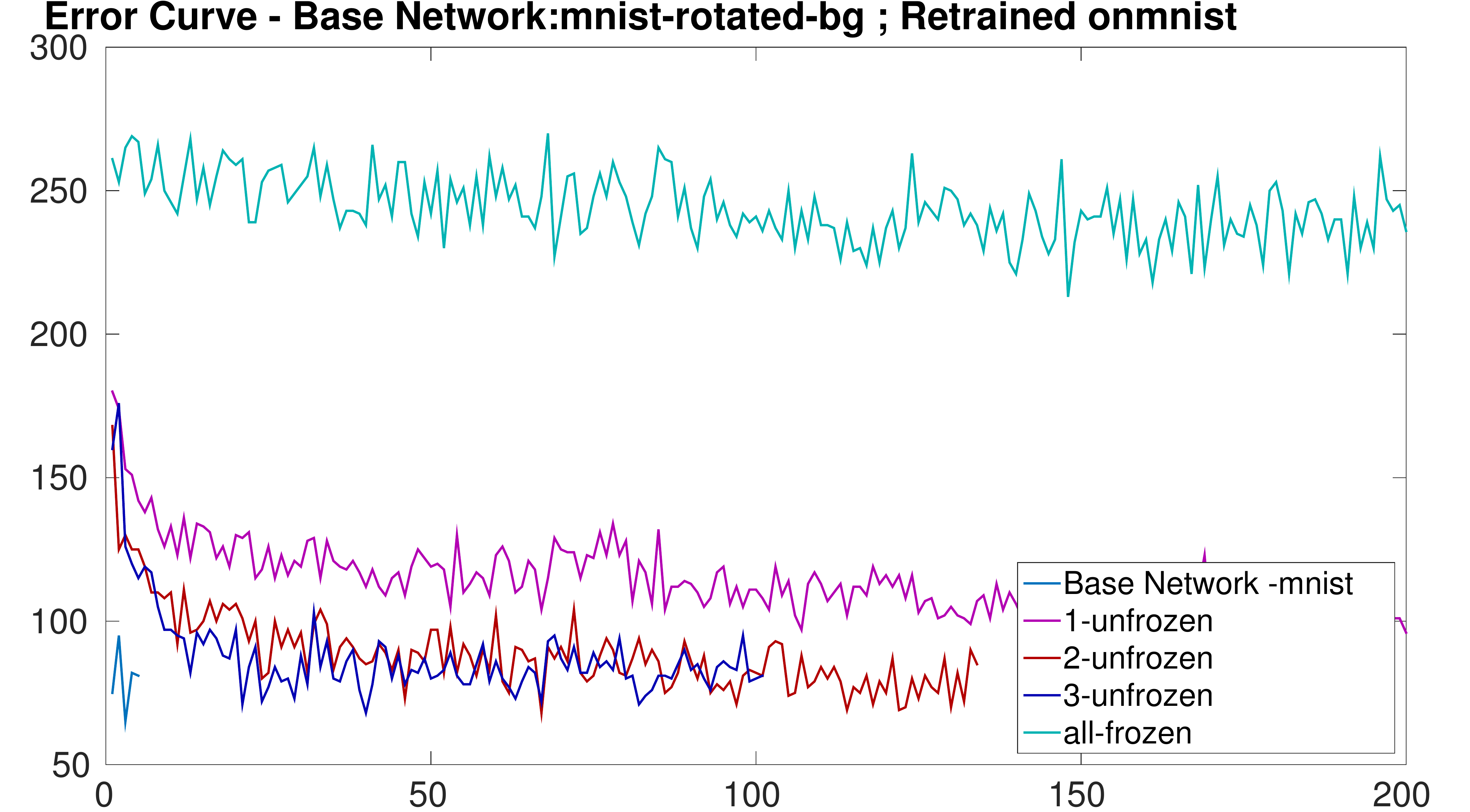} 		
	\end{center}
	\caption{Validation errors vs Epoch number for base-MNIST-rotated-bg retrained on MNIST}
	\label{fig:erros}
\end{figure}

Figure~\ref{fig:erros} demonstrates some interesting phenomenon that we discovered often. \emph{The gain in performance achieved, constantly decreases with increase in degrees of freedom.} Through the epochs, unfreezing only the classifier layer, quickly converges. But while unfreezing, all layers converge at about the same number of epochs. We also observe, that MNIST retrained over MNIST-rotated-background, with \emph{the last degree of freedom does not learn antything at all}. The error rate is within the statistical margin of error. This is a testament to the generality of MNIST-rotated-background among the MNIST datasets. One might expect this because MNIST-rotated-background contains smooth background images (similar to natural image set) and MNIST characters that are rotated. These conditions provide for a good generality.

\begin{table}[t]
	\begin{center}
		\begin{tabular}{||c| c 		|| 		c	 | 	c 	| 	c | 	c 	||}
			\hline
			$p$	 		& base 	& $k=0$      &  $k=1$      &   $k=2$  & $k=3$ \\							
			\hline \hline
			$1$				& Random 		& -		& -		& - 	& 55.61 \\
			& MNIST$[458]$ & 73.07 	& 73.91 & 76.37 & 77.52 \\
			\hline
			$3$				& Random 		& -		& -		& - 	& 73.34 \\
			& MNIST$[458]$ & 83.61 	& 87.2 	& 85.7 	& 87.6 \\
			\hline
			$5$				& Random 		& -		& -		& - 	& 83.32 \\
			& MNIST$[458]$ & 90.98 	& 92.98 & 92.6 	& 92.07 \\
			\hline	
			$10$			& Random 		& -		& -		& - 	& 81.31 \\
			& MNIST$[458]$ 	& 91.55 	& 93.71 & 93.82 & 95.08 \\
			\hline	
			$20$			& Random 		& -		& -		& - 	& 87.77 \\
			& MNIST$[458]$ 	& 95.52 	& 95.52 & 97.07 & 96.78 \\
			\hline	
			$30$			& Random 		& -		& -		& - 	& 88.62 \\
			& MNIST$[458]$ 	& 96.5			& 97.34	& 97.35 & 97.45 \\
			\hline			$50$			& Random 		& -		& -		& - 	& 90.78 \\
			& MNIST$[458]$ 	&  96.38		& 97.40		& 97.71		& 97.38\\
			\hline
			\hline
		\end{tabular}
	\end{center}
	\caption{Sub-sample experiment and its generalization accuracies for different layers of freezing. The re-train network was MNIST$[0,1,2,3,6,7,9]$. For obvious reasons random initializations are trained only with all layers unfrozen, hence the missing values.}
	\label{table:class-gen}
\end{table}

\begin{figure}[!t]
	\begin{center}
		\includegraphics[width=.99\linewidth]{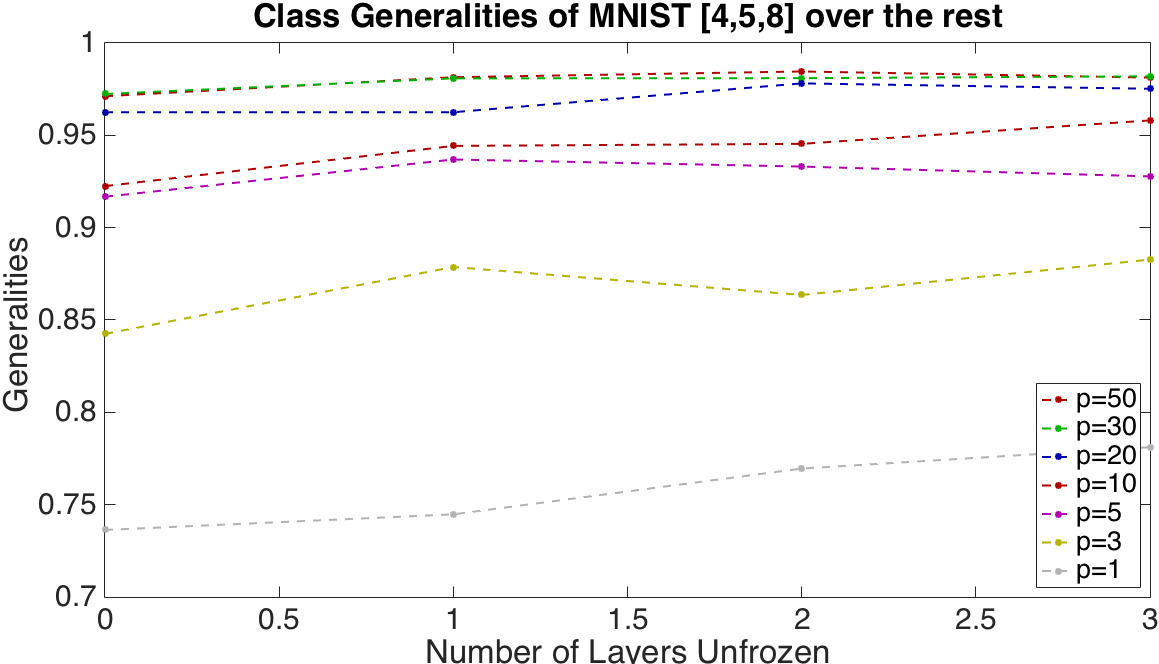} 		
	\end{center}
	\caption{Sub-class generalities for MNIST $[4,5,8]$}
	\label{fig:sub-gen}
\end{figure}

For the intra-class experiment described in subsection~\ref{sec:class-gen} above, table~\ref{table:class-gen} shows the accuracies. From the table one can observe that even with one-sample per class, a $7$-way classifier could achieve 22\% more accuracy than a randomly initialized network. It is note worthy that the last row of table~\ref{table:class-gen} still has $100$ times less data than the full dataset and it already achieves close to state-of-the-art accuracy even when no layer is allowed to change. This is a remarkably strong indicator that the classes $[4, 5, 8]$ generalizes the entire dataset. 

Figure~\ref{fig:sub-gen} mimics the same. We also observed that once initialized with a general enough subset of classes from within the same dataset, the generalities didn't vary among the layers like it did when we initialized with data from outside the mother dataset. We also observed that the more the data we used, more stable the generalities remained. Point of take away from this experiment is that if the classes are general enough, one may now initialize the network with only those classes and then learn the rest of the dataset even with very small number of samples.

\subsection{CIFAR 10 vs. Caltech 101}

From figure~\ref{fig:gen-results} we observe that Caltech 101 doesn't generalize to Cifar 10, which is surprising because Caltech 101 has a lot more classes. One would expect it to be more general. Its quite the opposite because Caltech 101 although has a lot of classes, the variability of each class is not as much as the variability in the Cifar 10 dataset.  But it is altogether a serendipitous result that \emph{Cifar 10 is more general than Caltech 101 on the lower layers}. However after three layers of obstination, we find that when the generalities crosses $1$, the effect nullifies and reverses slightly. Even though the low-level features are more general in Cifar 10, Caltech 101 generalizes more on higher layers.  

\subsection{Caltech 101 vs. Colonoscopy}

\begin{figure*}[!t]
	\begin{center}
		\includegraphics[width=0.24\linewidth]{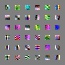} \rulesep	
		\includegraphics[width=0.24\linewidth]{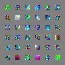} \rulesep		
		\includegraphics[width=0.24\linewidth]{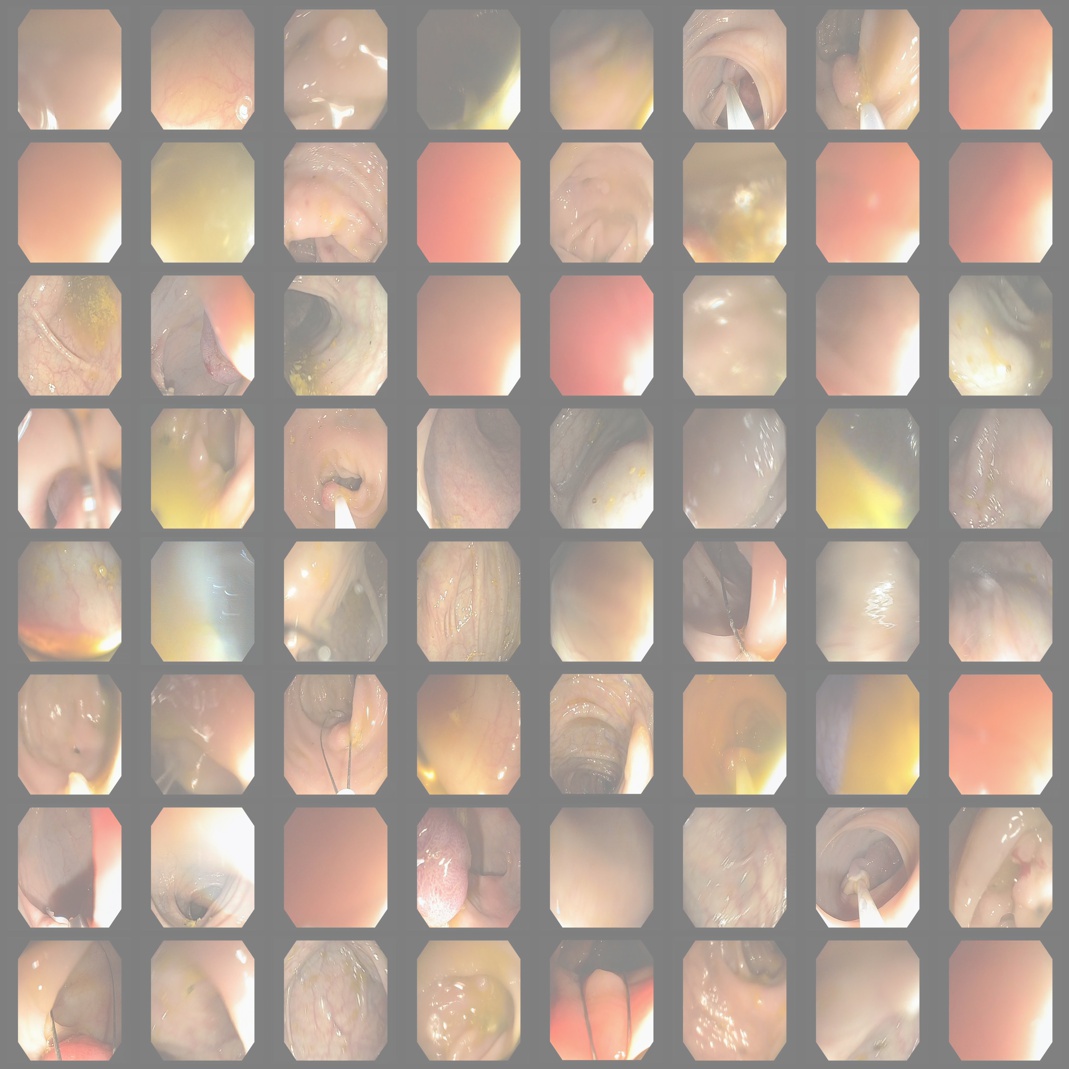}	\rulesep
		\includegraphics[width=0.24\linewidth]{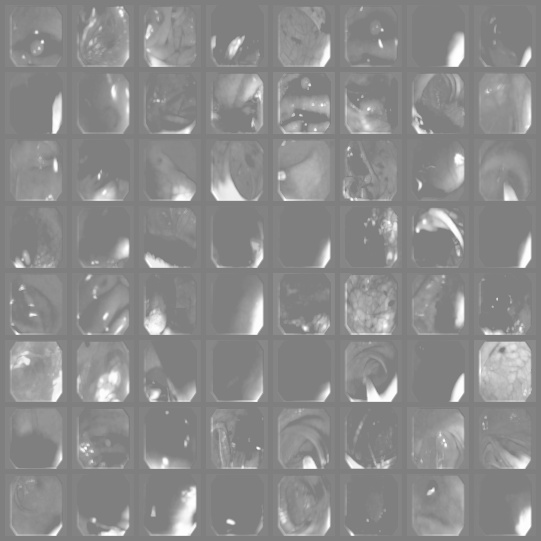}						
	\end{center}
	\caption{From left to right, separated by a line are filters learnt by a base Caltech101 base colonoscopy, sample images from the colonoscopy dataset and their first activation for a filter that detects smooth areas of brightness.}
	\label{fig:filters}
\end{figure*}

The colonoscopy dataset's labels identify if a image is deemed to be of a quality that is good enough so as to make a diagnosis on the pathology of that particular image. Figure~\ref{fig:filters} show the filters learnt by Caltech 101 base network and Colonoscopy base network for the exact same architecture from random initialization. Two things are immediately apparent from the learnt filters that while Caltech 101 learns more structured and organized shape features, Colonoscopy dataset learns at first sight what appears to be unstructured blob detectors and detectors for dark colors. These features still produce state-of-the-art accuracy on the dataset. On observation of the activations produced after the first layer, and from observations of images and their labels, one can immediately recognize that what the network is learning is indeed changes in brightness patterns.

Most often the video quality in colonoscopy is affected because of saturation when too much light is thrown at a scene. The quality is also affected due to light reflection from bodily fluids that is also noticeable in the activations. As also can be noticed that most of the filter colors are yellowish or blueish. On an colonoscopy video most often the video is also labelled poor quality when these colors are present, as these colors are often present mostly because of scattering and reflections. Having made these observations one would arrive at the obvious conclusion that neither dataset generalizes the other. This was indeed the result observed from figure~\ref{fig:gen-results}. Although, Caltech 101 seem to generalize a bit better for even though it predominantly learns shapes, it learns some color features also. 

\subsection{Summary of results}
From all these results and observations, we could summarize that one should prefer to initialize with a general dataset that might have a lot of variability or rather generality in data, when attempting to train with very few number of samples. Whenever possible one must initialize the network trained by a general dataset as this always boosts generalization performance. When there are biased datasets with large number of samples in some classes and fewer in others, one should train the most general classes first. Once the network is well-prejudiced one should start introducing the classes with fewer number of and less general samples, provided the general class is general enough.

\section{Conclusions}
\label{sec:conclusions}

In this paper, we used the performance of CNNs on a dataset when initialized with the filters from other datasets as a tool to measure generality. We proposed a generality metric using these generalization performances. We used the proposed metric to compare popular character recognition datasets and found some interesting patterns and generality assumptions that add to the knowledge-base of these datasets. In particular, we noticed that MNIST data is one of the most specific dataset. We also found that Char74k Kannada is less general than English datasets. We also calculated generality on class-level within a dataset and conclude that a few well-chosen classes used as pre-training could build a network that is well-initialized that even with $100$ times less samples, we could learn the other classes. We also provided some practical guidelines for a CNN engineer to adopt. After performing similar experiments on popular imaging datasets and medical datasets, we made similar serendipitous observations. 

{\small	
\bibliographystyle{ieee}
\bibliography{arxiv}
}

\end{document}